\documentclass{article}

\usepackage{arxiv}

\usepackage[utf8]{inputenc} 
\usepackage[T1]{fontenc}    
\usepackage{hyperref}       
\usepackage{url}            
\usepackage{booktabs}       
\usepackage{amsfonts}       
\usepackage{nicefrac}       
\usepackage{microtype}      
\usepackage{lipsum}		
\usepackage{graphicx}
\usepackage{doi}
\usepackage{algorithm}
\usepackage{algorithmic}
\usepackage{amsmath,amssymb,amsfonts}
\usepackage{graphicx}
\usepackage{textcomp}
\usepackage{caption} 
\usepackage{subcaption}
\usepackage{mathtools}
\usepackage{float}
\usepackage{booktabs,siunitx}
\usepackage{makecell}
\usepackage{multirow}
\usepackage{xcolor}

\usepackage{amsmath}

\title{SMU: smooth activation function for deep networks using smoothing maximum technique}

\author{
Koushik Biswas\\
	\And
	Sandeep Kumar\\
	\And
	 Shilpak Banerjee\\
	\And
	 Ashish Kumar Pandey\\
}

\date{}

\renewcommand{\shorttitle}{SMU: Smooth Maximum Unit}

\hypersetup{pdftitle={SMOOTH ACTIVATION FUNCTION},pdfsubject={Activation Function},pdfauthor={Koushik Biswas},pdfkeywords={Activation Function},}

\begin{document}
\maketitle

\begin{abstract}
Deep learning researchers have a keen interest in proposing two new novel activation functions which can boost network performance. A good choice of activation function can have significant consequences in improving network performance. A handcrafted activation is the most common choice in neural network models. ReLU is the most common choice in the deep learning community due to its simplicity though ReLU has some serious drawbacks. In this paper, we have proposed a new novel activation function based on approximation of known activation functions like Leaky ReLU, and we call this function Smooth Maximum Unit (SMU).  Replacing ReLU by SMU, we have got 6.22\% improvement in the CIFAR100 dataset with the ShuffleNet V2 model.
\end{abstract}


\section{Introduction}
Deep Neural network has emerged a lot in recent years and has significantly impacted our real-life applications. Neural networks are the backbone of deep learning. An activation function is the brain of the neural network, which plays a central role in the effectiveness \& training dynamics of deep neural networks. Hand-designed activation functions are quite a common choice in neural network models. ReLU \cite{relu} is a widely used hand-designed activation function. Despite its simplicity, ReLU has a major drawback, known as the dying ReLU problem in which up to 50\% neurons can be dead during network training. To overcome the shortcomings of ReLU, a significant number of activations have been proposed in recent years, and Leaky ReLU \cite{lrelu}, Parametric ReLU \cite{prelu}, ELU \cite{elu}, Softplus \cite{softplus}, Randomized Leaky ReLU \cite{rlrelu} are a few of them though they marginally improve performance of ReLU. Swish \cite{swish} is a non-linear activation function proposed by the Google brain team, and it shows some good improvement of ReLU. GELU \cite{gelu} is an another popular smooth activation function. It can be shown that Swish and GELU both are a smooth approximation of ReLU. Recently, a few non-linear activations have been proposed which improves the performance of ReLU, Swish or GELU. Some of them are either hand-designed or smooth approximation of Leaky ReLU function, and Mish \cite{mish}, ErfAct \cite{erfact}, Pad\'e activation unit \cite{pau}, Orthogonal Pad\'e activation unit \cite{opau} are a few of them.

\section{Related Work and Motivation}
In a deep neural network, activations are either fixed before training or trainable. Researchers have proposed several activations in recent years by combining known functions. Some of these functions have hyperparameters or trainable parameters. In the case of trainable activation functions, parameters are optimized during training. Swish is a popular activation function that can be used as either a constant or trainable activation function, and it shows some good performance in a variety of deep learning tasks like image classification, object detection, machine translation etc. GELU shares similar properties like the Swish activation function, and it gains popularity in the deep learning community due to its efficacy in natural language processing tasks. GELU has been used in BERT \cite{bert}, GPT-2 \cite{gpt2}, and GPT-3 \cite{gpt3} architectures. Pad\'e activation unit (PAU) has been proposed recently, and it is constructed from the approximation of the Leaky ReLU function by rational polynomials of a given order. Though PAU improves network performance in the image classification problem over ReLU, its variants, and Swish, it has a major drawback. PAU contains many trainable parameters, which significantly increases the network complexity and computational cost. 

Motivated from these works, we propose activation functions using the smoothing maximum technique. The maximum function is non-smooth at the origin. We want to explore how the smooth approximation of the maximum function (which can be used as an activation function) affects a network's training dynamics and performance. Our experimental evaluation shows that our proposed activation functions are comparatively more effective than ReLU, Mish, Swish, GELU, PAU etc., across different deep learning tasks. We summarise the paper as follows:
\begin{enumerate}
    \item We have proposed activation functions by smoothing the maximum function. We show that it can approximate GELU, ReLU, Leaky ReLU or the general Maxout family.
    \item We show that the proposed functions outperform widely used activation functions in a variety of deep learning tasks.
\end{enumerate}


\begin{figure*}[!t]
\begin{minipage}[t]{.305\linewidth}
        \centering
         \includegraphics[width=\linewidth]{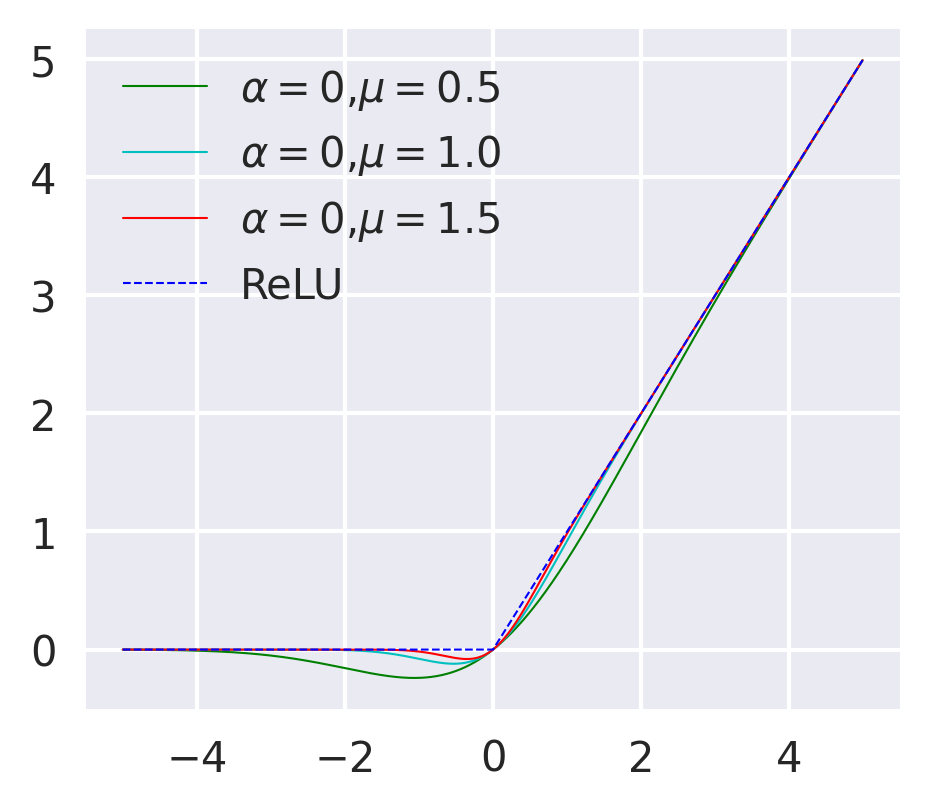}
        
        \caption{Approximation of ReLU using SMU ($\alpha = 0$) for different values of $\mu$. As $\mu \rightarrow \infty$, SMU smoothly approximate ReLU}
        \label{smu1}
         \end{minipage}
         \hfill
   \begin{minipage}[t]{.32\linewidth}
        \centering
    
         \includegraphics[width=\linewidth]{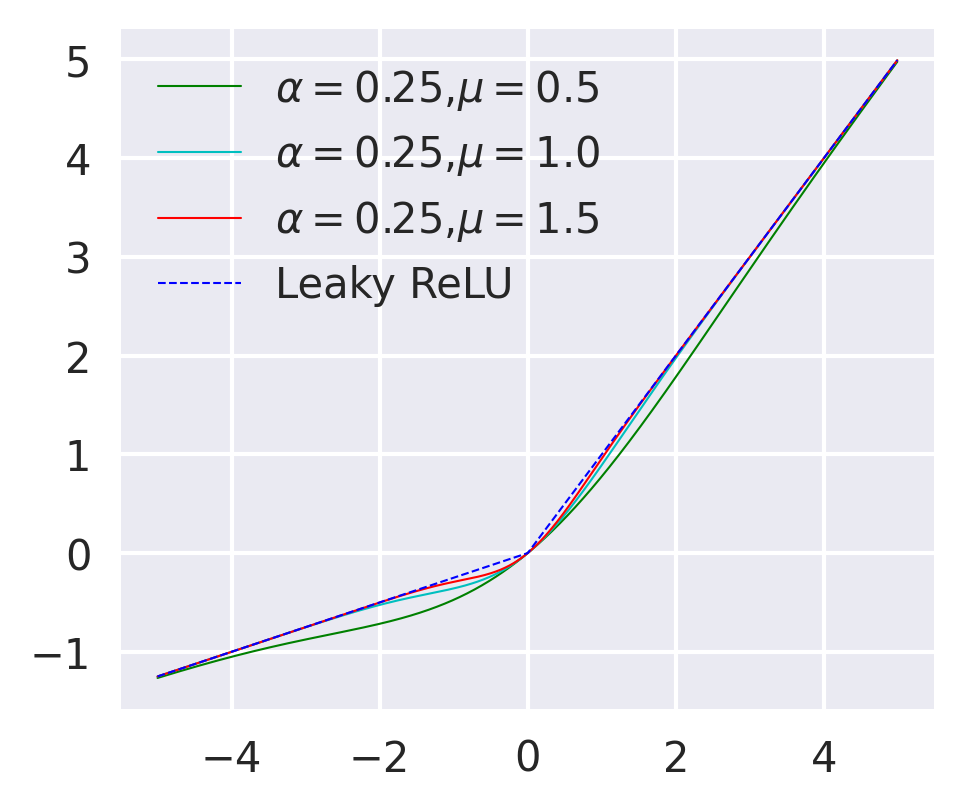}
        
        \caption{Approximation of Leaky ReLU ($\alpha = 0.25$) using SMU for different values of $\mu$. As $\mu \rightarrow \infty$, SMU smoothly approximate Leaky ReLU}
        \label{smu2}
   \end{minipage}
    \hfill
   \begin{minipage}[t]{.32\linewidth}
        \centering
    
         \includegraphics[width=\linewidth]{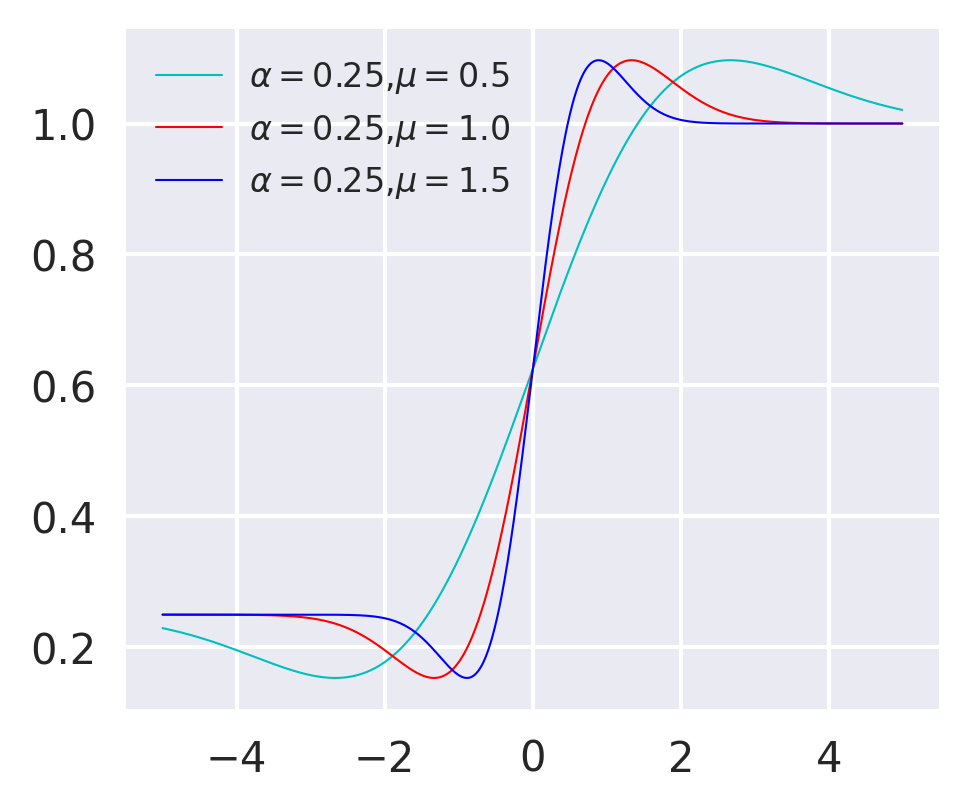}
       
        \caption{First order derivatives of SMU for $\alpha=0.25$ and different values of $\mu$.}
        \label{der}
   \end{minipage}
\end{figure*}

\section{Smooth Maximum Unit}
We present Smooth Maximum Unit (SMU), smooth activation functions from the smooth approximation of the maximum function. Using the smooth approximation of the $|x|$ function, one can find a general approximating formula for the maximum function, which can smoothly approximate the general Maxout\cite{maxout} family, ReLU, Leaky ReLU or its variants, Swish etc. We also show that the well established GELU \cite{gelu} function can be obtained as a special case of SMU.
\subsection{Smooth approximation of the maximum function}
Note that the maximum function can be expressed as following two different ways:
\begin{align}\label{eq1}
    max(x_1,x_2) &=\begin{cases} x_1 &\text{ if }x_1\geq x_2\\ x_2&\text{otherwise}\end{cases} \notag\\
    &=\frac{(x_1+x_2)+|x_1-x_2|}{2}
\end{align}
Note that the max function is not differentiable at the origin. Using approximations of the $|x|$ function by a smooth function, we can create approximations to the maximum functions. There are many known approximations to $|x|$, but for the rest of this article, we will focus on two specific approximations of $|x|$, namely $x\text{erf}(\mu x)$ and $\sqrt{x^2 + {\mu}^2}$. We noticed that the activations constructed using these two functions provide good performance on standard datasets on different deep learning problems (for more details, see the supplementary section). Note that $\sqrt{x^2 + {\mu}^2}$ as $\mu \rightarrow 0$ approximate $|x|$ from above while $x\text{erf}(\mu x)$. as $\mu \rightarrow \infty$ gives an approximation of $|x|$ from below. The approximation is uniform on compact subsets of the real line. Here $\operatorname{erf}$ is the Gaussian error function defined as follows:
\begin{align*}
    \operatorname{erf}(x) = \dfrac{2}{\sqrt{\pi}} \int_{0}^{x} e^{-t^2} \,dt. 
\end{align*}

Now, replacing the $|x|$ function by $x\text{erf}(\mu x)$ in equation (\ref{eq1}), we have the smooth approximation formula for maximum function as follows:

\begin{align}\label{eq2}
    f_1(x_1,x_2;\mu) = \frac{(x_1+x_2)+(x_1-x_2)\ \text{erf}(\mu(x_1-x_2))}{2}.
\end{align}
Similarly, we can derive the the smooth approximation formula for the maximum function from equation (\ref{eq1}) by replacing the $|x|$ function by $\sqrt{x^2 + {\mu}^2}$ as follows:
\begin{align}\label{eq3}
    f_2(x_1,x_2;\mu) = \frac{(x_1+x_2)+\sqrt{(x_1-x_2)^2+\mu^2}}{2}
\end{align}
Note that as $\mu \rightarrow \infty$, $f_1(x_1,x_2;\mu)\rightarrow$ max$(x_1,x_2)$ and as $\mu \rightarrow 0$, $f_2(x_1,x_2;\mu)\rightarrow$ max$(x_1,x_2)$.
For particular values of $x_1$ and $x_2$, we can approximate known activation functions. For example, consider $x_1=ax$, $x_2=bx$, with $a\neq b$ in (\ref{eq2}), we get:
\begin{align}\label{eq4}
    f_1(ax,bx;\mu) = \frac{(a+b)x+(a-b)x\ \text{erf}(\mu(a-b)x)}{2}.
\end{align}
This is a simple case from the Maxout family \cite{maxout} while more complicated cases can be found by considering nonlinear choices of $x_1$ and $x_2$. We can similarly get smooth approximations to ReLU and Leaky ReLU. For example, consider $x_1=x$ and $x_2=0$, we have smooth approximation of ReLU as follows:
\begin{align}\label{eq5}
     f_1(x,0;\mu) = \frac{x+x\ \text{erf}(\mu x)}{2}.
\end{align}
We know that GELU \cite{gelu} is a smooth approximation of ReLU. Notice that, if we choose $\mu = \frac{1}{\sqrt{2}}$ in equation (\ref{eq5}), we can recover GELU activation function which also show that GELU is smooth approximation of ReLU. Also, considering $x_1=x$ and $x_2=\alpha x$, we have a smooth approximation of Leaky ReLU or Parametric ReLU depending on whether $\alpha$ is a hyperparameter or a learnable parameter.

\begin{align}\label{eq6}
     f_1(x,\alpha x ;\mu) = \frac{(1+\alpha)x+(1-\alpha)x\ \text{erf}(\mu(1-\alpha)x)}{2}.
\end{align}
Note that, equation (\ref{eq5}) and equation (\ref{eq6}) approximate ReLU or Leaky ReLU from below. Similarly, we can derive approximating function from equation (\ref{eq3}) which will approximate ReLU or Leaky ReLU from above.

The  corresponding  derivatives  of  equation (\ref{eq6}) for input variable $x$ is 
\begin{align}\label{eq7}
    \frac{d}{dx}f_1(x,\alpha x ;\mu)=\frac{1}{2}[(1+\alpha)+(1-\alpha)\ \text{erf}(\mu(1-\alpha)x)\notag \\
    +\frac{2}{\sqrt{\pi}}\mu(1-\alpha)^2xe^{-(\mu(1-\alpha)x)^2}]\\
\text{where}\ \  \frac{d}{dx}
    \text{erf}(x)=\frac{2}{\sqrt{\pi}}e^{-x^2}.\quad\quad\quad\quad\quad\quad\quad\quad\quad\quad\quad\notag
\end{align}
Figures~\ref{smu1}, ~\ref{smu2}, and ~\ref{der} show the plots for $f_1(x,0; \mu)$, $f_1(x,0.25x; \mu)$, and derivative of $f_1(x,0.25x; \mu)$ for different values of $\mu$. From the figures it is clear that as $\mu \rightarrow \infty$, $f_1(x,\alpha x;\mu)$ smoothly approximate ReLU or Leaky ReLU depending on value of $\alpha$. We call the function in equation~(\ref{eq6}) as Smooth Maximum Unit (SMU). Similarly, We can derive a function by replacing $x_1=x$ and $x_2=\alpha x$ in equation~(\ref{eq3}) and we call this function SMU-1. For all of our experiments, we will use SMU and SMU-1 as our proposed activation functions. 

\subsection{Learning activation parameters via back-propagation} Trainable activation function parameters are updated using backpropagation \cite{backp} technique. We implemented forward pass in both Pytorch \cite{pytorch} \& Tensorflow-Keras \cite{keras} API, and automatic differentiation will update the parameters. Alternatively, CUDA \cite{cuda} based implementation (see \cite{lrelu}) can be used and the gradients of equations \ref{eq8} and \ref{eq9}) for the  parameters $\alpha$ and $\mu$ can be computed as follows:
\begin{align}\label{eq8}
    \frac{\partial f_1}{\partial \alpha} = \frac{1}{2}[x-x\ \text{erf}(\mu(1-\alpha)x)-(1-\alpha)\mu x^2e^{-(\mu(1-\alpha)x)^2}] 
    \end{align}
    \begin{align}\label{eq9}
\frac{\partial f_1}{\partial \mu} = \frac{1}{2}(1-\alpha)^2x^2e^{-(\mu(1-\alpha)x)^2}
 \end{align}

$\alpha$ and $\mu$ can be either hyperparameters or trainable parameters. For classification problem, we consider $\alpha=0.25$, a hyperparameter. We consider $\mu$ as trainable parameter and initialise at $1.0$ and $4.352665993287951e^{-09}$ for SMU and SMU-1 respectively. For object detection and semantic segmentation problem, we consider $\alpha=0.01$, a hyperparameter. We consider $\mu$ as trainable parameter and initialise at $2.5$ and $4.332461424154261e-09$ for SMU and SMU-1 respectively. \\
Now, note that the class of neural networks with SMU and SMU-1 activation function is dense in $C(K)$, where $K$ is a compact subset of $\mathbb{R}^n$ and $C(K)$ is the space of all continuous functions over $K$.\\
The proof follows from the following proposition (see \cite{pau}). 

\textbf{Proposition 1. (Theorem 1.1 in Kidger and Lyons, 2020 \cite{universal}) :-} Let $\rho: \mathbb{R}\rightarrow \mathbb{R}$ be any continuous function. Let $N_n^{\rho}$ represent the class of neural networks with activation function $\rho$, with $n$ neurons in the input layer, one neuron in the output layer, and one hidden layer with an arbitrary number of neurons. Let $K \subseteq \mathbb{R}^n$ be compact. Then $N_n^{\rho}$ is dense in $C(K)$ if and only if $\rho$ is non-polynomial.

\section{Experiments}
In the next subsections, we will provide a detailed experimental evaluation to compare performance for our proposed activation function and some widely used activation functions on 3 different deep learning problems like image classification, object detection, and semantic segmentation. 
\subsection{CIFAR}\label{cifardata}
We report the top-1 accuracy on Table~\ref{tab2} and Table~\ref{tab3} on CIFAR10 \cite{cifar10} and CIFAR100 \cite{cifar10} datasets respectively. We consider MobileNet V2 \cite{mobile}, Shufflenet V2 \cite{shufflenet}, PreActResNet \cite{preactresnet}, ResNet \cite{resnet}, VGG \cite{vgg} (with batch-normalization \cite{batch}), and EfficientNet B0 \cite{efficientnet}.



\begin{table}[!htbp]
\makebox[\textwidth][c]{
    \begin{tabular}{c|c|c|c}
\toprule
        Model    &\multicolumn{1}{c|}{ReLU} & \multicolumn{1}{c|}{SMU} & \multicolumn{1}{c}{SMU-1} \\
\midrule
 &  Top-1 accuracy   & Top-1 accuracy & Top-1 accuracy \\
\midrule

ShuffleNet V2 1.0x & 90.81 $\pm$ 0.24 & \textbf{92.72} $\pm$ 0.18 & \textbf{92.42} $\pm$ 0.20\\

ShuffleNet V2 2.0x & 91.70 $\pm$ 0.20 & \textbf{93.61} $\pm$ 0.14
& \textbf{93.40} $\pm$ 0.16\\
\midrule 

PreActResNet 34 & 94.21 $\pm$ 0.17 & \textbf{95.12} $\pm$ 0.13& \textbf{94.93} $\pm$ 0.14\\

\midrule
ResNet 34 & 94.22 $\pm$ 0.18 & \textbf{94.91} $\pm$ 0.16& \textbf{94.77} $\pm$ 0.17\\
\midrule
SeNet 34 & 94.42 $\pm$ 0.20 & \textbf{95.27} $\pm$ 0.15 & \textbf{94.89} $\pm$ 0.17\\
\midrule
MobileNet V2 & 94.22 $\pm$ 0.15  &   \textbf{95.50} $\pm$ 0.09&   \textbf{95.27} $\pm$ 0.10\\
\midrule
EffitientNet B0 &  95.10 $\pm$ 0.15 &  \textbf{96.23} $\pm$ 0.10 &  \textbf{96.11} $\pm$ 0.12 \\
\midrule
Xception & 90.51 $\pm$ 0.22 & \textbf{93.25} $\pm$ 0.17 & \textbf{92.59} $\pm$ 0.20\\
\midrule
VGG16 & 93.59 $\pm$ 0.18 & \textbf{94.54} $\pm$ 0.14 & \textbf{93.32} $\pm$ 0.15\\
\bottomrule
\end{tabular}%
}
\vspace{0.2cm}
\caption{Comparison between our proposed activations and other baseline activations in the CIFAR10 dataset. We report the results for the mean of 15 different runs. mean$\pm$std is reported in the table.}
\label{tab2}%
\end{table}%



\begin{table}[!htbp]
\makebox[\textwidth][c]{
    \begin{tabular}{c|c|c|c}
\toprule
        Model    &\multicolumn{1}{c|}{ReLU} & \multicolumn{1}{c|}{SMU} & \multicolumn{1}{c}{SMU-1} \\
\midrule
 &  Top-1 accuracy   & Top-1 accuracy & Top-1 accuracy \\
\midrule

Shufflenet V2 1.0x & 64.41 $\pm$ 0.25 & \textbf{70.60} $\pm$ 0.21 & \textbf{69.96} $\pm$ 0.22\\
Shufflenet V2 2.0x & 67.52 $\pm$ 0.25 & \textbf{73.74} $\pm$ 0.20
& \textbf{73.45} $\pm$ 0.23\\
\midrule 

PreActResNet 34 & 73.41 $\pm$ 0.24 & \textbf{76.21} $\pm$  0.20& \textbf{75.87} $\pm$  0.21\\

\midrule

ResNet 34 & 73.33 $\pm$ 0.27 & \textbf{75.77} $\pm$ 0.20& \textbf{75.59} $\pm$ 0.21\\

\midrule

SeNet 34 & 75.12 $\pm$ 0.22 & \textbf{76.79} $\pm$ 0.18 & \textbf{75.79} $\pm$ 0.21\\
\midrule

MobileNet V2 & 74.17 $\pm$ 0.24  & \textbf{76.31} $\pm$ 0.19  & \textbf{76.03} $\pm$ 0.19  \\

\midrule

Xception  & 71.22 $\pm$ 0.26  & \textbf{74.62} $\pm$ 0.23  & \textbf{74.11} $\pm$ 0.23  \\
\midrule
EffitientNet B0 & 76.60 $\pm$ 0.27  &  \textbf{79.10} $\pm$ 0.22&  \textbf{78.77} $\pm$ 0.23 \\
\midrule
VGG16 & 71.87  $\pm$ 0.30 & \textbf{73.26} $\pm$ 0.22& \textbf{72.79} $\pm$ 0.23\\
\bottomrule
\end{tabular}%
}
\vspace{0.2cm}
\caption{Comparison between our proposed activations and other baseline activations in the CIFAR100 dataset. We report the results for the mean of 15 different runs. mean$\pm$std is reported in the table.}
\label{tab3}
\end{table}


\subsection{Object Detection}
In this section, we report results on Pascal VOC dataset \cite{pascal} on Single Shot MultiBox Detector(SSD) 300 \cite{ssd} with VGG-16(with batch-normalization) \cite{vgg} as the backbone network. We report the mean average precision (mAP) in Table~\ref{tabod} for the mean of 6 different runs.

\begin{table}[!htbp]
\begin{center}
\begin{tabular}{ |c|c|c| }
 \hline
 Activation Function &  \makecell{mAP}  \\
  \hline
SMU &  \textbf{78.1}$\pm$0.09 \\ 
 \hline
 SMU-1 &  \textbf{77.9}$\pm$0.10 \\ 
 \hline
 ReLU  &  77.2$\pm$0.14  \\ 
 \hline
 
 Leaky ReLU &  77.2$\pm$0.18 \\
 \hline
 PReLU & 77.2$\pm$0.17\\
\hline
ReLU6 & 77.1$\pm$0.15\\

\hline
 ELU  &  75.1$\pm$0.22\\
 \hline
 Softplus & 74.2$\pm$0.25\\
 \hline
Swish  &  77.3$\pm$0.11\\
 \hline
 GELU & 77.3$\pm$0.12\\
 \hline

 \end{tabular}
 \vspace{0.2cm}
\caption{A detailed comparison between SMU activation and other baseline activations in Pascal VOC dataset. mean$\pm$std is reported in the table} 
\label{tabod}
\end{center}
\end{table}
\subsection{Semantic Segmentation}
We report our experimental results in this section on the popular CityScapes dataset \cite{city}. The U-net model \cite{unet} is considered as the segmentation framework, and we report the mean of 6 different runs for Pixel Accuracy and the mean Intersection-Over-Union (mIOU) on test data on table~\ref{tabsg}.
\begin{table}[!htbp]
\begin{center}
\begin{tabular}{ |c|c|c|c| }
 \hline
 Activation Function & \makecell{
Pixel\\ Accuracy} & mIOU  \\
 \hline
 SMU & \textbf{81.71}$\pm$0.38 & \textbf{71.45}$\pm$0.30 \\ 
 \hline
  SMU-1 & \textbf{81.17}$\pm$0.41 & \textbf{71.10}$\pm$0.30 \\ 
 \hline
 ReLU  & 79.49$\pm$0.46 & 69.31$\pm$0.28 \\ 
 \hline
  
 PReLU & 78.95$\pm$0.42 & 68.88$\pm$0.41\\
 \hline
 ReLU6 & 79.58$\pm$0.41 & 69.70$\pm$0.42\\
 \hline
  Leaky ReLU & 79.41$\pm$0.41 & 69.64$\pm$0.42\\
 \hline

 ELU  & 79.48$\pm$0.50 & 68.19$\pm$0.40\\
 \hline
 Softplus & 78.45$\pm$0.52 & 68.08$\pm$0.49\\
 \hline
  Swish  & 80.22$\pm$0.46 & 69.81$\pm$0.30\\
 \hline

 GELU & 80.14$\pm$0.37 & 69.59$\pm$0.40\\

 \hline
 \end{tabular}
 \vspace{0.3cm}
\caption{A detailed comparison between SMU activation and other baseline activations in CityScapes dataset. mean$\pm$std is reported in the table.} 
\label{tabsg}
\end{center}
\end{table}

\section{Conclusion}
This work uses the maximum smoothing technique to approximate Leaky ReLU, a well-established non-smooth (not differentiable at 0) function by two smooth functions. We call these two functions SMU and SMU-1, and we use them as a potential candidate for activation functions. Our experimental evaluation shows that the proposed functions beat the traditional activation functions in well-known deep learning problems and have the potential to replace them.   
\bibliography{references.bib}
\bibliographystyle{unsrt}

\end{document}